\documentclass[12pt,onecolumn]{IEEEtran}
\usepackage{cite}
\usepackage{amsmath,amssymb,amsfonts}
\usepackage{algorithmic}
\usepackage{graphicx}
\usepackage{textcomp}
\usepackage{xcolor}
\usepackage{tabularx} 
\def\BibTeX{{\rm B\kern-.05em{\sc i\kern-.025em b}\kern-.08em
		T\kern-.1667em\lower.7ex\hbox{E}\kern-.125emX}}

\begin{document}
	
	\title{QYOLO: Lightweight Object Detection via Quantum Inspired Shared Channel Mixing}
	
	\author{\IEEEauthorblockN{Garvit Kumar Mittal}\\
		\IEEEauthorblockA{\textit{Central Research Laboratory}
			\textit{Bharat Electronics Limited}
			Ghaziabad, India \\
			gkmb2543@gmail.com}\\
		\and
    \IEEEauthorblockN{Sahil Tomar}\\
		\IEEEauthorblockA{\textit{Central Research Laboratory}
	\textit{Bharat Electronics Limited}
	Ghaziabad, India \\
			sahiltomar893@gmail.com}\\
		\and
		\IEEEauthorblockN{Sandeep Kumar}\\
		\IEEEauthorblockA{\textit{Central Research Laboratory}
	\textit{Bharat Electronics Limited}
	Ghaziabad, India \\
			sann.kaushik@gmail.com}\\
	}
	
	\maketitle
	
	\begin{abstract}
	The rapid advancement of object detection architectures has positioned single-stage detectors as the dominant solution for real-time visual perception. A primary source of computational overhead in these models lies in the deep backbone stages, where C2f bottleneck modules at high stride levels accumulate a disproportionate share of parameters due to quadratic scaling with channel width.
This work introduces {QYOLO}, a quantum-inspired channel mixing framework that achieves genuine architectural compression by replacing the two deepest backbone C2f modules at $P4/16$ (512 channels) and $P5/32$ (1024 channels) with a compact {QMixBlock}. The proposed block performs global channel recalibration through a sinusoidal mixing mechanism with shared learnable parameters across both backbone stages, enforcing consistent channel importance without requiring independent per-stage parameter sets.
The neck and detection head remain fully classical and unchanged. Evaluation on the VisDrone2019 benchmark demonstrates that QYOLOv8n achieves a $20.2\%$ reduction in parameter count (3.01M to 2.40M) and $12.3\%$ GFLOPs reduction with only $0.4$ pp mAP@50 degradation. QYOLOv8s achieves $21.8\%$ reduction with $0.1$ pp degradation.
When combined with knowledge distillation, full accuracy parity is recovered at no cost to compression. An expanded backbone-plus-neck variant achieved $38$--$41\%$ reduction at the cost of greater accuracy degradation, motivating the backbone-only final design.

	\end{abstract}
	

\section{Introduction}

The widespread adoption of convolutional neural networks has positioned object detection as a core component of modern autonomous systems, supporting applications such as unmanned aerial platforms, robotic perception, and continuous visual monitoring. Among real-time detection frameworks, single-stage paradigms have become the preferred solution due to their favorable latency characteristics. Recent detection architectures have achieved notable gains in accuracy through increasingly elaborate backbone and feature aggregation strategies \cite{ref1}. Despite these advances, the associated computational cost has grown substantially, limiting practical deployment on hardware-constrained edge devices.

A primary source of this parameter overhead lies in the deep backbone stages of single-stage detectors. Architectures such as YOLOv8 employ a convolutional backbone in which C2f bottleneck modules at the highest stride levels - $P4/16$ (512 channels) and $P5/32$ (1024 channels) - accumulate a disproportionate share of total parameters due to quadratic growth in parameter count with channel width. These stages encode high-level semantic features and operate at compressed spatial resolutions, yet they remain implemented with local $3 \times 3$ spatial convolutions that impose significant memory and compute costs. Subsequent neck and head components that perform multi-scale fusion are architecturally important but computationally lighter; addressing redundancy at the backbone level therefore offers greater leverage for model compression without disrupting spatial fusion logic.

Existing model compression strategies address this problem from the outside rather than through architectural redesign. Pruning methods fall into two fundamentally different categories with distinct implications for deployment. Unstructured (weight-level) pruning \cite{ref2} zeroes individual weight entries based on magnitude, producing sparse tensors without modifying the network graph; the stored parameter count is unchanged and inference acceleration requires sparse execution hardware or libraries that most edge platforms do not support \cite{ref3}. True structured pruning removes entire filters or channels, producing a smaller dense network that works on standard hardware without sparse libraries \cite{ref3}. A prior work [5] demonstrates this approach on YOLOv8, reducing parameters from 25.85M to 6.85M through a three-stage pipeline of sparsity-aware training, channel pruning, and knowledge distillation, albeit with a non-trivial accuracy drop that necessitates distillation for recovery. In the experiments reported here, the Pruned-20 baseline applies magnitude-based weight sparsification to YOLOv8n at $20\%$ sparsity, the resulting model retains 3.01M parameters identical to the unpruned baseline, confirming that no structural channel elimination was performed in this configuration.

Knowledge distillation (KD)\cite{ref5} can transfer representational capacity from a large teacher to a smaller student, but it neither guarantees a smaller model nor avoids the added training complexity of multi-stage supervision. Channel recalibration approaches such as squeeze-and-excitation \cite{ref6} and efficient channel attention \cite{ref7} dynamically adjust feature responses but assign independent parameter sets to each scale, limiting their ability to capture globally consistent channel relationships and offering no reduction in the underlying parameter count.

Recent developments in quantum machine learning offer a complementary perspective on efficient representation learning. The formulation of parameterized quantum circuits \cite{ref8} demonstrates that periodic transformations with entangled parameter structures can encode complex distributions with comparatively few degrees of freedom. Early work on quantum neural networks established the theoretical feasibility of learning-based quantum models \cite{ref9}. Related findings in implicit neural representations \cite{ref10} further show that sinusoidal activation functions - mathematically analogous to quantum rotation operations - capture high-frequency detail more efficiently than piecewise linear nonlinearities such as ReLU or SiLU. This spectral advantage is particularly relevant for the VisDrone2019 benchmark \cite{ref11}, where targets frequently occupy fewer than $32 \times 32$ pixels and strong spatial cues are absent, placing a premium on representational efficiency per parameter.

Motivated by these observations, this work presents {QYOLO}, a parameter-efficient object detection framework that introduces a Quantum-Inspired Channel Mixing module ({QMixBlock}) as a direct architectural replacement for the deep backbone C2f bottlenecks of YOLOv8. The QMixBlock computes channel importance weights via a compact sinusoidal function, and uses them to recalibrate backbone feature maps at the $P4$ and $P5$ stride levels. The entire neck, detection head, and early backbone stages remain unchanged.

A design variant extended QMixBlock replacement into three neck fusion layers in addition to the two backbone stages, achieving $38$--$41\%$ parameter reduction but at a measurable accuracy cost and requiring substantially longer training to converge. The final backbone-only design achieves approximately $20\%$ parameter reduction with only $0.4$ percentage points of mAP@50 degradation - a more practical trade-off for edge deployment.

The novel contributions of this work are:

\begin{itemize}
    \item {Backbone-Targeted Quantum-Inspired Mixing:} The C2f bottleneck modules at backbone layers 6 ($P4/16$, 512 channels) and 8 ($P5/32$, 1024 channels) are replaced with QMixBlock, a sinusoidal channel recalibration module with shared learnable parameters.
    This surgical backbone-only modification leaves the neck and detection head fully intact, concentrating compression at the highest-parameter backbone stages.

    \item {Shared Parameter Mixing Across Backbone Stages:} A single SharedQuantumMixer module - comprising learnable weight vector $w$ and phase offset $\Theta$ - is instantiated once and shared across both backbone QMixBlock instances. This enforces consistent channel importance patterns between the $P4$ and $P5$ semantic feature stages, acting as an implicit regulariser while eliminating redundant per-stage parameter sets.

    \item {Genuine Architectural Compression with Empirical Validation:} Unlike pruning-based methods that preserve the original parameter count while introducing weight sparsity, QMixBlock achieves genuine architectural parameter elimination. QYOLOv8n reduces parameters from 3.01M to 2.40M ($20.2\%$ reduction) and GFLOPs from 8.1 to 7.1 with only $0.4$ pp mAP@50 degradation on VisDrone2019. QYOLOv8s achieves $21.8\%$ reduction with $0.1$ pp degradation. When combined with knowledge distillation, full accuracy parity is recovered at no cost to the compression ratio.
\end{itemize}
\section{Proposed Methodology}

This section presents the conceptual motivation, mathematical foundation, and architectural realisation of the Quantum-Inspired Channel Mixing (QMix) framework. The methodology targets the long-standing trade-off between computational efficiency and representational capacity in single-stage object detection by replacing the two most parameter-intensive backbone bottleneck modules with a compact sinusoidal channel recalibration design grounded in principles drawn from quantum information theory.

\begin{figure}[t]
\centering
\includegraphics[width=\columnwidth]{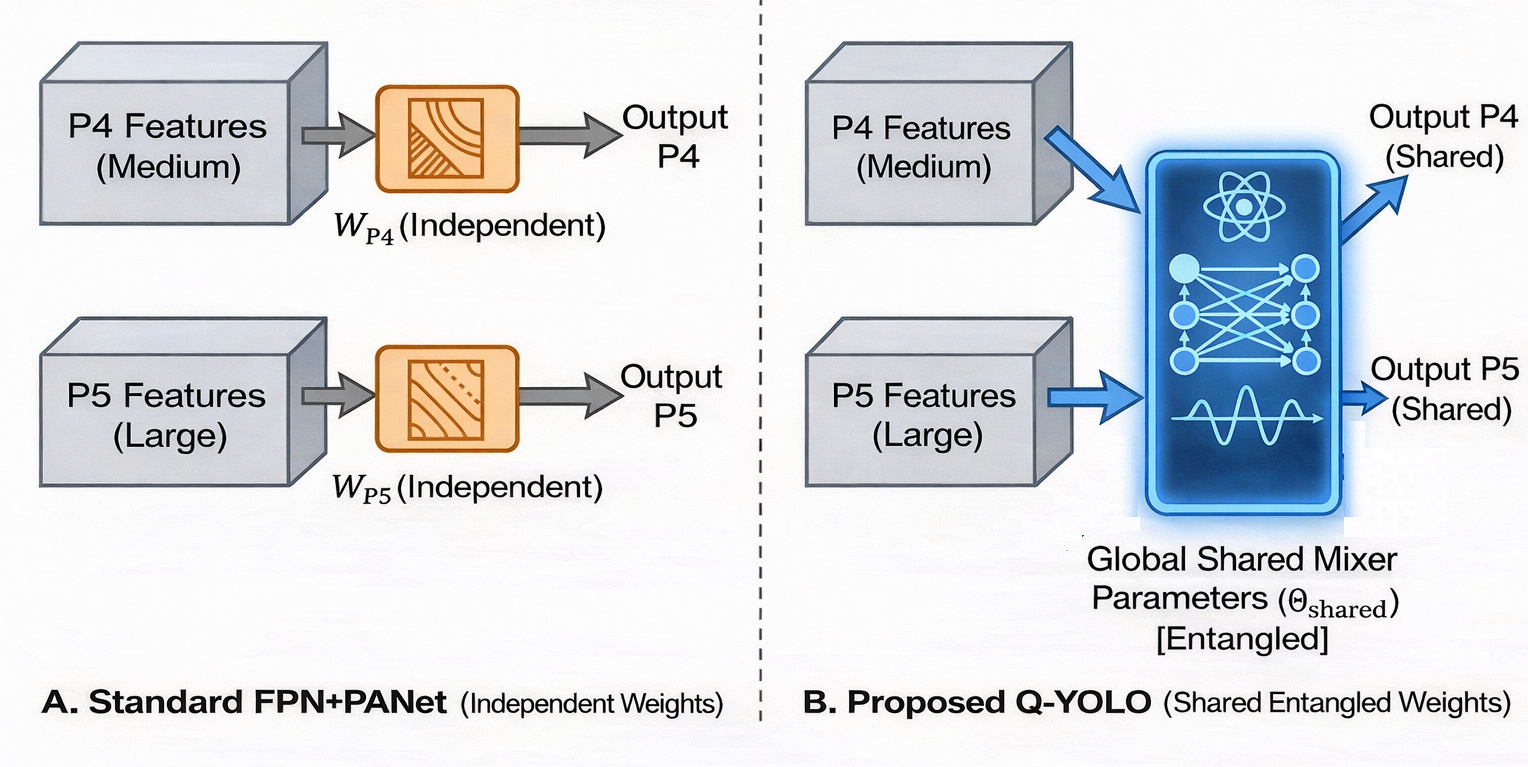}
\caption{Conceptual Comparison of Parameter Sharing Strategies in Feature Pyramid Networks}
\label{fig:shared_mixture}
\end{figure}

Fig.~\ref{fig:qyolo_arch} provides an overview of the proposed QYOLO architecture, illustrating the placement of QMixBlock modules at backbone layers 6 (P4/16) and 8 (P5/32) and internal processing stages of shared mixer, and Fig. ~\ref{fig:shared_mixture} shows the conceptual comparison of parameter sharing strategies in FPN. The proposed approach departs from redundancy-intensive local spatial convolution at the deepest backbone stages and instead adopts a globally coordinated channel mixing mechanism with strong spectral expressiveness, leaving the neck and detection head architecturally unchanged.

\begin{figure*}[t]
\centering
\includegraphics[width=\textwidth]{qmix_architecture.png}
\caption{Overview of the proposed QYOLO architecture. The QMixBlock replaces C2f modules at backbone layers P4/16 and P5/32. The shared quantum-inspired mixer performs sinusoidal channel recalibration using global context aggregation, latent projection, and periodic mixing.}
\label{fig:qyolo_arch}
\end{figure*}

\subsection{Computational Limitations of Deep Backbone Bottlenecks}

A clear understanding of the proposed design requires examination of the structural constraints present in widely adopted single-stage detection pipelines. Architectures such as YOLOv8 employ a convolutional backbone to extract hierarchical spatial representations, followed by a Feature Pyramid Network (FPN) combined with a Path Aggregation Network (PAN) neck \cite{ref12, ref13} for multi-scale fusion, and a detection head for prediction. While the neck handles spatial feature aggregation and is architecturally critical, the deepest backbone stages - the C2f modules at stride $P4/16$ (512 channels) and stride $P5/32$ (1024 channels) - accumulate the greatest share of total parameters because parameter cost scales quadratically with channel width at these positions.

\subsubsection{Redundancy in Deep Backbone Convolutional Bottlenecks}

At backbone layers 6 and 8, C2f modules apply repeated $3 \times 3$ convolutional bottleneck transformations over the highest-channel-count feature maps in the network. For an input feature tensor $X \in \mathbb{R}^{C \times H \times W}$, each such transformation is expressed as:
\begin{equation}
F_{\text{bottleneck}}(X) = \phi(W_2 * \phi(W_1 * X)) + X
\end{equation}
where $*$ denotes convolution, $W_1$ and $W_2$ are learnable kernels of spatial size $3 \times 3$, and $\phi(\cdot)$ is a monotonic activation function such as SiLU. The parameter count of each convolutional layer scales as:
\begin{equation}
N_{\text{params}} \propto \mathcal{O}(C_{\text{in}} \cdot C_{\text{out}} \cdot K^2)
\end{equation}
At $P4/16$ ($C = 512$) and $P5/32$ ($C = 1024$) in the small configuration, and their width-scaled equivalents in the nano configuration, this quadratic dependence concentrates a disproportionate fraction of total model parameters in precisely these two backbone stages. Empirical compression studies indicate that a significant fraction of these weights contribute minimally to detection performance, reflecting substantial redundancy in local convolutional processing at depth.

\subsubsection{Limitations of Local Receptive Fields at Backbone Depth}

Standard $3 \times 3$ convolutions operate within a restricted spatial neighbourhood and are limited in their ability to encode global channel-level relationships. At the deep backbone stages $P4$ and $P5$, the primary objective shifts from local edge detection to high-level semantic feature encoding - a role where global channel interactions carry more discriminative information than local spatial filtering. Capturing such relationships through stacked local convolutions requires many layers and incurs unnecessary parameter cost. A globally shared, non-local channel recalibration mechanism offers a more parameter-efficient alternative for this semantic encoding role.

\subsection{Quantum-Classical Correspondence}

The central idea of the proposed framework is the reinterpretation of quantum mechanical principles within a classical neural architecture. Two principles are of particular relevance: unitary rotation and parameter entanglement. These concepts are drawn from parameterised quantum circuits, also referred to as variational quantum circuits \cite{ref8}. Early work on quantum neural networks established the theoretical feasibility of learning-based quantum models \cite{ref9}.

\subsubsection{Quantum-Inspired Neuron Representation}

In quantum computation, information is encoded within the state vector $|\psi\rangle$ of a qubit. A fundamental operation is a unitary rotation \cite{ref8} about the $Y$ axis of the Bloch sphere:
\begin{equation}
R_y(\theta) = \exp\left(-i\frac{\theta}{2}\hat{Y}\right) = I \cos\left(\frac{\theta}{2}\right) - i \hat{Y} \sin\left(\frac{\theta}{2}\right)
\end{equation}
where $\hat{Y}$ denotes the Pauli-$Y$ operator, $I$ is the identity matrix, and $\theta$ is a tunable parameter. This highlights a defining property of quantum evolution: periodicity. The output state oscillates smoothly as $\theta$ varies, enabling rich nonlinear behaviour with compact parameterisation. This motivates adopting sinusoidal transformations in the proposed channel mixing mechanism - periodic functions that provide expressive feature modulation without reliance on deep or wide convolutional stacks.

\subsubsection{Spectral Bias and Periodic Activation Functions}

Classical deep networks based on monotonic activations such as ReLU, SiLU, or Tanh exhibit spectral bias \cite{ref14}: they preferentially learn low-frequency components during training, while encoding high-frequency patterns requires substantial increases in network depth or width \cite{ref14}. This limitation is particularly pronounced in aerial imagery benchmarks such as VisDrone \cite{ref11}, where targets frequently occupy fewer than $32 \times 32$ pixels and high-frequency boundary cues carry critical discriminative information. Sinusoidal activations provide oscillatory responses that naturally encode high-frequency variation within a compact parameter space. A single sinusoidal unit can represent patterns that would otherwise require many piecewise linear units, increasing representational density per parameter.

\subsubsection{Parameter Sharing Through Entanglement}

In quantum physics, entanglement links the state of one particle intrinsically to another independent of spatial separation. This principle is translated into the backbone design by treating the two deep backbone stages as components of a unified feature extraction system. In the baseline architecture, layers 6 and 8 maintain independent C2f parameter sets $\Phi_{P4}$ and $\Phi_{P5}$, implicitly assuming that channel importance logic differs fundamentally between depth levels. The proposed framework challenges this assumption: channel relationships that discriminate foreground structure from background noise are expected to remain broadly consistent across $P4$ and $P5$, since both encode high-level semantic features differing primarily in spatial scale rather than semantic category. A single shared mixing parameter set $\{w, \Theta\}$ is therefore applied across both backbone QMixBlock instances, enforcing consistent channel importance while acting as an implicit regulariser.

\subsection{Mathematical Formulation of the QMixBlock}

The QMixBlock is defined as a modular, fully differentiable replacement for the C2f bottleneck modules at backbone layers 6 ($P4/16$) and 8 ($P5/32$). Its internal computation is structured into five sequential stages.

\subsubsection{Stage 1: Global Context Aggregation}

Convolutional operators are effective for local spatial patterns but limited in their ability to capture global semantic context. To obtain a compact global channel descriptor, global average pooling is applied. Given an input feature tensor $X \in \mathbb{R}^{B \times C \times H \times W}$, the channel descriptor $g \in \mathbb{R}^{B \times C}$ is computed as:
\begin{equation}
g_c = \frac{1}{H W} \sum_i \sum_j X_{c,i,j}
\end{equation}
This collapses the spatial dimensions while preserving global per-channel statistics, providing a compact representation of each channel's global activation pattern.

\subsubsection{Stage 2: Manifold Projection}

The channel descriptor $g$ is projected into a lower-dimensional latent space using a linear compression matrix $W_{\text{compress}} \in \mathbb{R}^{r \times C}$, where the latent r is controlled by reduction ratio $R$:
\begin{equation}
r = \frac{C}{R}, \quad z = W_{\text{compress}} \cdot g
\end{equation}
This reduces the computational burden of subsequent processing while retaining salient semantic information. The projection preserves the most discriminative channel relationships while significantly reducing the dimensionality over which sinusoidal mixing operates.

\subsubsection{Stage 3: Quantum-Inspired Mixing}

The latent vector $z$ is processed by the SharedQuantumMixer - the core computational unit of the QMixBlock. Rather than a conventional multilayer perceptron, the mixer applies a single sinusoidal transformation with learnable per-dimension weighting and phase offset in equation 6, the frequency of the sinusoidal function is fixed at unity, representational expressiveness arises from the per-dimension weighting $w$ and adaptive phase alignment $\Theta$ rather than from a learnable global frequency scalar. The mixing H is a single-layer operation, applied in one pass over the $r$-dimensional latent space.
    \begin{equation}
H = \sin(z \cdot w + \Theta)
\end{equation}
This formulation provides several advantages. The sinusoidal function has non-zero derivatives of all orders, enabling complex curvature modelling in the latent space. The weight vector $w$ introduces per-dimension frequency sensitivity, while $\Theta$ enables adaptive alignment of periodic responses to feature distributions. All computations remain real-valued, ensuring compatibility with standard acceleration hardware.

\subsubsection{Stage 4: Channel Excitation}

The mixed latent representation $H$ is projected back to the original channel dimension using a linear expansion matrix $W_{\text{expand}} \in \mathbb{R}^{C \times r}$, followed by sigmoid gating to produce normalised channel importance weights:
\begin{equation}
s = \sigma(W_{\text{expand}} \cdot H)
\end{equation}
where $s \in \mathbb{R}^{B \times C}$ encodes a soft importance score for each channel in the range $(0, 1)$.

\subsubsection{Stage 5: Feature Recalibration}

The original feature tensor is rescaled using the computed channel weights through element-wise multiplication:
\begin{equation}
Y_{\text{out}} = X \odot s
\end{equation}
This selectively amplifies informative channels while suppressing less relevant responses, providing global context-driven backbone feature refinement without any local spatial convolution. The output $Y_{\text{out}}$ retains the same spatial dimensions as the input $X$, but with recalibrated per-channel activations.

\subsection{Global Entanglement and Shared Parameterisation}

A central contribution of the framework is the shared parameter strategy applied across the two backbone QMixBlock instances. In the standard YOLOv8 backbone, each of layers 6 and 8 maintains independent C2f parameter sets $\Phi_{P4}$ and $\Phi_{P5}$. Such independence implicitly assumes that channel importance logic differs fundamentally between backbone depth levels. The proposed approach challenges this assumption: channel relationships that discriminate foreground structure from background noise are expected to remain broadly consistent across $P4$ and $P5$, since both encode high-level semantic features that differ primarily in spatial scale rather than semantic category.

Under the proposed formulation, a single SharedQuantumMixer module is instantiated with parameter set:
\begin{equation}
\Theta_{\text{shared}} = \{w, \Theta\}, \quad w \in \mathbb{R}^{1 \times r}, \quad \Theta \in \mathbb{R}^{1 \times r}
\end{equation}
This shared parameter set governs the sinusoidal mixing at both backbone QMixBlock instances. During forward propagation, the latent representations from each instance are processed by the same transformation:
\begin{align}
z_{P4} &\rightarrow H_{P4} = \sin(z_{P4} \cdot w + \Theta) \\
z_{P5} &\rightarrow H_{P5} = \sin(z_{P5} \cdot w + \Theta)
\end{align}
where the same $w$ and $\Theta$ governed by $\Theta_{\text{shared}}$ are applied at each stage. The neck - which includes C2f modules at layers 12, 15, 18, and 21 - does not participate in this shared mechanism and retains fully independent classical C2f parameters throughout.

This shared parameterisation encourages the learning of depth-consistent channel interactions across $P4$ and $P5$, acts as an implicit regulariser by preventing unnecessary divergence between the two backbone levels, and reduces the total parameter count by sharing what would otherwise be two independent sets of mixer parameters.

\subsection{Complexity and Efficiency Analysis}

To quantify the efficiency benefit of replacing backbone C2f modules with QMixBlock, a parameter complexity comparison is conducted.

\subsubsection{Baseline C2f Bottleneck Complexity}

A standard C2f backbone module at channel width $C$ (assuming $C_{in}$ = $C_{out}$ = C) consists of repeated $3 \times 3$ convolutional bottlenecks. For each bottleneck, the parameter count scales quadratically:
\begin{equation}
N_{\text{baseline}} \approx 9C^2
\end{equation}
At $P4/16$ ($C = 512$) and $P5/32$ ($C = 1024$) in the small model configuration, this results in several million parameters concentrated at just two backbone positions.

\subsubsection{QMixBlock Complexity}

The QMixBlock replaces spatial convolutions with linear projection and element-wise sinusoidal mixing. Parameter cost is dominated by the compression and expansion projection matrices. The SharedQuantumMixer contributes parameters $w$ and $\Theta$, each of dimension $r$, which is negligible relative to the projection matrices.

The total parameter count per QMixBlock instance is:
\begin{equation}
N_{\text{QMix}} = C \cdot r + r \cdot C + 2r = \frac{2C^2}{R} + \frac{2C}{R}
\end{equation}
Since $\frac{2C}{R} \ll \frac{2C^2}{R}$ for practical values of $C$, the dominant term is:
\begin{equation}
N_{\text{QMix}} \approx \frac{2C^2}{R}
\end{equation}
For a reduction ratio of $R = 4$, as used in experiments:
\begin{equation}
N_{\text{QMix}} \approx 0.5C^2
\end{equation}
\subsubsection{Efficiency Ratio}

The per-module parameter reduction factor between a single C2f bottleneck and a QMixBlock at $R = 4$ is:
\begin{equation}
\frac{N_{\text{baseline}}}{N_{\text{QMix}}} \approx \frac{9C^2}{0.5C^2} = 18
\end{equation}
This approximately $18\times$ reduction per replaced module at the layer level, applied at two backbone positions, yields the approximately $20\%$ total model parameter reduction observed experimentally: QYOLOv8n (3.01M $\rightarrow$ 2.40M) and QYOLOv8s (11.13M $\rightarrow$ 8.70M). The remaining approximately $80\%$ of parameters in early backbone layers, the neck, and the detection head are entirely unchanged.

Because $w$ and $\Theta$ are shared across both QMixBlock instances rather than duplicated, the SharedQuantumMixer contributes only $2r$ parameters in total to the full model regardless of the number of QMixBlock placements.

\subsection{Implementation}

The proposed framework is implemented as a modular extension within the Ultralytics training environment (Ultralytics 8.4.26, PyTorch 2.5.1, Python 3.10.20, CUDA 12.1). A single SharedQuantumMixer module stores the learnable parameters $w$ and $\Theta$, both initialised as $w \sim \mathcal{N}(0,1)$ and $\Theta = 0$, allocated at dimension $\texttt{dim} = 1024$ with dynamic slicing to the actual latent size $r$ at runtime. Individual QMixBlock instances at backbone layers 6 and 8 reference this shared module by pointer during forward propagation, ensuring that gradients from both instances flow into the same $w$ and $\Theta$ parameters during backpropagation.

Integration is achieved by replacing the C2f module entries at layers 6 and 8 in the YOLOv8 YAML architecture configuration file with QMixBlock entries, while all remaining architecture definitions remain unchanged. No custom CUDA kernels or specialised compiler operations are required. All five stages of the QMixBlock - global average pooling, linear compression, sinusoidal transformation, sigmoid gating, and element-wise recalibration - rely exclusively on standard PyTorch primitives supported by automatic differentiation.

This design ensures compatibility with common deployment toolchains, including ONNX, TensorRT, and CoreML, without requiring specialised hardware beyond general-purpose GPU accelerators.
\section{Experimental Setup}

This section describes the experimental design used to evaluate the effectiveness of the proposed QYOLO framework. The evaluation focuses on model size, computational cost, and detection accuracy. Details regarding the dataset, baseline models, training configurations, hardware environment, and evaluation criteria are provided to ensure transparency and reproducibility.
\subsection{Dataset Description}

Experiments are conducted on the VisDrone2019-DET benchmark \cite{ref11}, a large-scale aerial object detection dataset. It is chosen for its challenging characteristics, including small object sizes (often $<32 \times 32$ pixels), high object density with frequent occlusions, and class imbalance across categories.

The dataset contains 10,209 images captured under diverse conditions. Following the standard split, 6,471 images are used for training, 548 for validation, and 3,190 for testing. Annotations cover ten object classes, including pedestrian, car, bicycle, and bus.

\subsection{Baseline Architectures}

The YOLOv8 Nano (YOLOv8n) and YOLOv8 Small (YOLOv8s) configurations are selected as primary baselines \cite{ref1}, covering the spectrum from extreme edge constraints to moderately capable embedded platforms. YOLOv8 introduces the C2f bottleneck module, which enhances gradient propagation through increased feature splitting. Despite this improvement, the deepest C2f modules at backbone layers 6 and 8 contribute substantial parameter overhead due to quadratic scaling with channel width at $P4/16$ (512 channels) and $P5/32$ (1024 channels).

For all primary QYOLO configurations, QMixBlock replaces C2f exclusively at backbone layers 6 and 8. All neck and detection head components remain identical to the standard YOLOv8 configuration. An earlier design variant (QYOLO-v0) additionally replaces three neck C2f modules, achieving greater compression at a measurable accuracy cost; this is reported as an upper-bound compression reference.

\subsection{Implementation and Training Protocols}

All models are implemented using PyTorch 2.5.1 (CUDA 12.1), Python 3.10.20, and the Ultralytics framework version 8.4.26. Primary baseline and QMixBlock variants are trained from random initialisation under identical conditions. 
\subsection{Training Configuration}

\noindent
{Training configuration used across all experiments.}

\begin{center}
\begin{tabularx}{\columnwidth}{lX}
\hline
{Component} & {Setting} \\
\hline
Optimizer & AdamW (momentum=0.937, weight decay=0.0005) \\
Learning Rate & 0.001 $\rightarrow$ 0.00001 (linear decay) \\
Batch Size & 16 \\
Epochs & 300 \\
Input Resolution & $640 \times 640$ \\
\hline
\end{tabularx}
\end{center}

\subsection{Data Augmentation}

\noindent
{Standard augmentation techniques are applied as follows:}

\begin{center}
\begin{tabular}{ll}
\hline
{Method} & {Details} \\
\hline
Mosaic & Enabled (probability = 1.0) \\
MixUp & Probability = 0.15 \\
Geometric & Scaling, translation, horizontal flip \\
\hline
\end{tabular}
\end{center}
\subsection{Hardware Environment}

Experiments are conducted on a workstation with NVIDIA RTX 6000 Ada GPUs (49 GB VRAM), an Intel Xeon W9-3595X CPU, and 512 GB RAM.

\subsection{Evaluation Metrics}

Performance is evaluated using standard detection metrics: 
{Parameter Count} (model size), 
{GFLOPs} (computational complexity), 
{mAP@50} (primary detection accuracy), and 
{mAP@50--95} (strict localization performance across IoU thresholds).

\section{Results and Discussion}

This section presents an analysis of the experimental results obtained from evaluating the proposed QYOLO framework on the VisDrone2019 DET benchmark. The discussion is organised as follows. Section IV-A reports the primary results and outlines the design progression from QYOLO-v0 to the final architecture. Section IV-B provides a per-class performance analysis. Section IV-C presents an ablation study to isolate the impact of individual design components. Finally, Section IV-D contextualises the results by comparing QYOLO with alternative model compression strategies.

\subsection{Primary Results and Design Progression}

Table~\ref{tab:main_results} reports the primary detection results alongside the v0 baselines, enabling direct comparison between the two architectural configurations explored during development.

\begin{table*}[t]
\centering
\caption{Primary Results and Design Progression for QYOLOv8 Variants vs YOLOv8n on VisDrone2019 Validation Split}
\label{tab:main_results}
\begin{tabular}{lcccccccc}
\hline
Model & QMixBlocks & Params & Param Red. & GFLOPs & Train Time & mAP@50 & mAP@50--95 \\
\hline
YOLOv8n & 0 & 3.01M & -- & 8.1 & 4.44h & 34.9 & 20.1 \\
QYOLOv8n (v0) & 5 & 1.86M & 38.2\% & 6.2 & 5.76h & 32.6 & 18.5 \\
\textbf{QYOLOv8n} & \textbf{2} & \textbf{2.40M} & \textbf{20.2\%} & \textbf{7.1} & \textbf{3.46h} & \textbf{34.5} & \textbf{19.9} \\\hline
YOLOv8s & 0 & 11.13M & -- & 28.5 & 4.47h & 40.0 & 23.6 \\
QYOLOv8s (v0) & 5 & 6.54M & 41.2\% & 21.0 & 3.56h & 39.7 & 23.2 \\
\textbf{QYOLOv8s} & \textbf{2} & \textbf{8.70M} & \textbf{21.8\%} & \textbf{24.6} & \textbf{3.94h} & \textbf{39.9} & \textbf{23.4} \\\hline
\end{tabular}
\end{table*}

Primary QMixBlock and baseline runs were trained for a fixed 300-epoch budget without early stopping. In contrast, v0 models employed early stopping ($\text{patience}=100$), resulting in extended training durations (nano-v0: 446 epochs, small-v0: 263 epochs), indicating increased optimisation difficulty under aggressive compression.

QYOLOv8n achieves a $20.2\%$ reduction in parameter count (3.01M $\rightarrow$ 2.40M) and a $12.3\%$ reduction in GFLOPs (8.1 $\rightarrow$ 7.1), with only a marginal $0.4$ pp drop in mAP@50 and $0.2$ pp in mAP@50--95. These results are obtained under identical training conditions, confirming that performance differences arise solely from the architectural replacement of backbone C2f modules with QMixBlock. Similar trends are observed for YOLOv8s, demonstrating consistent gains across model scales.

Training efficiency also improves, with QYOLOv8n achieving approximately $22\%$ faster per-epoch training due to reduced backbone complexity.

\begin{table}[t]
\centering
\caption{Layer-wise QMixBlock Placement Across Variants}
\label{tab:placement}
\begin{tabular}{lccc}
\hline
Layer & Role & Baseline & QYOLO-v0 / Ours \\
\hline
\textbf{Layer 6 (Backbone)} & \textbf{P4/16} & \textbf{C2f} & \textbf{QMixBlock} \\
\textbf{Layer 8 (Backbone)} & \textbf{P5/32} & \textbf{C2f} & \textbf{QMixBlock} \\
Layer 12 (Neck) & P4 Top-down & C2f & QMixBlock (v0 only) \\
Layer 18 (Neck) & P4 Bottom-up & C2f & QMixBlock (v0 only) \\
Layer 21 (Neck) & P5 Bottom-up & C2f & QMixBlock (v0 only) \\
\hline
\end{tabular}
\end{table}

The v0 design extends QMixBlock replacement into the neck while preserving the P3 layer to maintain small-object detection performance. This configuration achieves higher compression (38--41\%) but introduces accuracy degradation and significantly harder optimisation.

These results highlight a key architectural insight: replacing backbone C2f modules is effective because these stages primarily encode high-level semantic information, where channel relationships dominate. In contrast, neck layers perform spatial feature fusion and rely on local convolutional structure; replacing them with global channel mixing disrupts spatial integration. 
The final backbone-only design therefore achieves a more favourable accuracy--compression trade-off by targeting parameter-heavy layers without compromising the spatial reasoning capabilities of the detection pipeline.

\subsection{Per-Class Analysis}

Table~\ref{tab:per_class} reports per-class AP@50 for YOLOv8n and QYOLOv8n on the VisDrone2019 validation set.
\begin{table}[t]
\centering
\caption{Per-Class AP@50 (\%) of YOLOv8n vs QYOLOv8n}
\label{tab:per_class}
\begin{tabular}{lccc}
\hline
Class & Baseline & QYOLO & $\Delta$ (pp) \\
\hline
Pedestrian & 39.1 & 37.4 & -1.7 \\
People & 30.2 & 30.3 & +0.1 \\
Bicycle & 8.9 & 9.1 & +0.2 \\
Car & 77.6 & 77.3 & -0.3 \\
\textbf{Van} & \textbf{37.7} & \textbf{39.1} & \textbf{+1.4} \\Truck & 29.0 & 28.5 & -0.5 \\
Tricycle & 24.4 & 22.4 & -2.0 \\
Awning-tricycle & 12.7 & 11.9 & -0.8 \\
\textbf{Bus} & \textbf{47.9} & \textbf{49.3} & \textbf{+1.4} \\Motor & 41.0 & 40.0 & -1.0 \\
\hline
ALL & 34.9 & 34.5 & -0.4 \\
\hline
\end{tabular}
\end{table}
The overall $-0.4$ pp change in mAP@50 reflects a balanced aggregation of class-wise gains and losses. Notably, \textit{van} and \textit{bus} improve by $+1.4$ pp, suggesting that categories with strong global appearance characteristics benefit from the sinusoidal channel mixing mechanism. Classes such as \textit{bicycle} and \textit{people} show negligible variation ($\leq 0.2$ pp), indicating minimal impact.

The largest regressions occur for \textit{tricycle} ($-2.0$ pp) and \textit{pedestrian} ($-1.7$ pp), both of which rely heavily on fine-grained spatial cues. These categories often involve small, densely packed, or partially occluded objects, where precise spatial boundary information is critical. The global average pooling stage in QMixBlock compresses spatial information into a channel descriptor, limiting its ability to preserve such fine details. A similar trend is observed for the \textit{motor} class.

These results indicate that QMixBlock is most effective for categories characterised by strong global feature signatures, while performance degrades for classes that depend on high-resolution spatial discrimination.

\subsection{Ablation Study}
\begin{table}[t] \centering \small \caption{Ablation Study of Design Variants of QMixBlock (YOLOv8n)} \label{tab:ablation} \resizebox{\columnwidth}{!}{ \begin{tabular}{lcccccc} \hline Variant & Formula & $\alpha$ & $\Theta$ & Params & mAP@50-95 \\ \hline QMix-Sin & $\sin(z \cdot w)$ & Fixed & Removed & 2.40M & 19.4 \\ QMix-Scaled & $\sin(\alpha z \cdot w + \Theta)$ & Learnable & Learnable & 2.40M & 19.5 \\ QMix-Full & $\sin(z \cdot w + \Theta)$ & Fixed & Learnable & 3.29M & 19.6 \\ \textbf{QMixBlock} & $\mathbf{\sin(z \cdot w + \Theta)}$ & \textbf{Fixed} & \textbf{Learnable} & \textbf{2.40M} & \textbf{19.9} \\ \hline \end{tabular} } \end{table}

{Role of the phase offset $\Theta$.} Removing the phase term (QMix-Sin) leads to a noticeable performance drop ($-0.5$ mAP), demonstrating that learnable phase alignment is essential for adapting to diverse channel-wise feature distributions. The phase term enables flexible feature shifting, which improves representational capacity without increasing parameter count.

{Role of frequency $\alpha$.} While introducing a learnable frequency (QMix-Scaled) slightly improves early-stage performance, it does not translate into better final accuracy. This suggests that additional frequency flexibility increases optimisation difficulty and may introduce training instability. In contrast, a fixed unit frequency acts as an implicit regulariser, promoting more stable convergence.

{Role of spatial convolution.} Incorporating a $3 \times 3$ spatial convolution (QMix-Full) provides marginal accuracy gains but incurs a $ 37\% $ increase in parameters. Although this branch enhances local spatial modelling, it conflicts with the lightweight design objective and offers limited efficiency–accuracy trade-off benefits.

The final QMixBlock achieves the best balance between accuracy and efficiency, outperforming all variants without increasing parameter count. The results indicate that phase adaptability is the most critical component, while additional frequency learning and spatial augmentation introduce unnecessary complexity with limited practical gains

\subsection{Comparison with other Compression Methods}

\setlength{\tabcolsep}{3pt} 
\begin{table}[t]
\centering
\footnotesize
\setlength{\tabcolsep}{4pt}
\renewcommand{\arraystretch}{0.95}

\caption{Comparison with Alternative Compression Strategies of YOLOv8n}
\label{tab:comparison}

\begin{tabular}{lcccc}
\hline
Method & Parameters & Reduction & mAP@50 \\
\hline
YOLOv8n Baseline  & 3.01M & -- & 34.9 \\
Unstructured Pruning (20\%) & 3.01M$^{*}$ & 0\% & 33.8 \\
QYOLOv8n & 2.40M & 20.2\% & 34.5 \\
QYOLOv8n + KD  & 2.40M$^{\dagger}$ & 20.2\% & 34.9 \\
\hline
\end{tabular}

*Pruning zeroes individual weights without altering the network graph, the
stored parameter count is therefore identical to the unpruned baseline.

† Knowledge Distillation uses QYOLOv8n as the student and the full YOLOv8n as the
teacher; the parameter count of the deployed model is that of the student
alone
\end{table}

Table ~\ref{tab:comparison} compares QYOLOv8n against direct compression baselines on the VisDrone2019 validation set. The pruning baseline applies unstructured sparsification at 20\%, zeroing individual weights without altering the network structure, resulting in no reduction in stored parameters (3.01M) and a 1.1 pp drop in mAP@50 \cite{ref3}. Moreover, such approaches require specialised sparse-execution support to yield runtime benefits, limiting their practicality on standard edge hardware.

In contrast, QYOLOv8n achieves a true 20.2\% parameter reduction with only a 0.4 pp decrease in mAP@50 through architectural compression, producing a smaller dense model directly compatible with conventional deployment frameworks. When combined with KD \cite{ref5} using YOLOv8n \cite{ref1} as the teacher, the accuracy gap is fully recovered, albeit with additional training complexity.

Compared to alternative lightweight detection strategies, channel-pruning methods \cite{ref3,ref15} and structured pruning pipelines \cite{ref4} achieve strong compression but require multi-stage optimisation and careful tuning. Accuracy-focused variants such as DCS-YOLOv8 \cite{ref16} improve detection performance at the cost of increased model size and latency, reducing their suitability for edge deployment. Two-stage detectors like Faster R-CNN \cite{ref17} and their edge adaptations \cite{ref18} remain computationally intensive and struggle to meet real-time constraints.
Overall, QYOLOv8n offers a favorable balance between efficiency and accuracy, it delivers genuine compression in a single training stage while maintaining near-baseline performance, making it well-suited for real-time aerial edge applications.

\section{Conclusion}
This work introduced QYOLO, a quantum-inspired architectural framework for efficient real-time object detection. The proposed design replaces the deepest backbone C2f bottlenecks with a shared QMixBlock, defined by sinusoidal function, applied at layers $P4/16$ and $P5/32$ while leaving the neck and detection head unchanged.
QYOLOv8n achieves a $20.2\%$ reduction in parameters and $12.3\%$ reduction in GFLOPs with only a $0.4$ pp drop in mAP@50, while QYOLOv8s demonstrates similar gains with negligible accuracy loss. Reduced backbone complexity also improves training efficiency, yielding faster per-epoch training.
Ablation studies confirm that the learnable phase offset $\Theta$ is essential for expressiveness, while a fixed unit frequency is more stable than a learnable alternative. More aggressive variants that extend QMixBlock into the neck achieve higher compression but introduce notable accuracy degradation, highlighting the effectiveness of the backbone-only design.

Unlike pruning-based approaches that rely on sparsity, QMixBlock provides genuine architectural compression by reducing parameters directly. Compared to knowledge distillation, which requires additional training complexity, QYOLO achieves near-parity accuracy in a single-stage pipeline, making it well-suited for deployment.
Future work will explore selective extension of QMixBlock into the neck and evaluate performance under low-precision inference.

\end{document}